\documentclass[twocolumn,10pt]{asme2ej}

\usepackage{graphicx} 
\usepackage{hyperref}   
\hypersetup{
	colorlinks=true,
	linkcolor=blue,
	citecolor=blue,
	urlcolor=blue,
}
\usepackage{amsmath}

\usepackage{algorithm}
\usepackage{algpseudocode}
\usepackage{amsmath}
\newcommand{\clip}{\mathrm{clip}}
\newcommand{\Unif}{\mathcal{U}}
\newcommand{\Normal}{\mathcal{N}}
\usepackage{booktabs}
\usepackage{tabularx}
\usepackage{array}
\usepackage{ragged2e}
\usepackage{rotating} 
\newcolumntype{Y}{>{\RaggedRight\arraybackslash}X}
\usepackage[square,numbers]{natbib}
\usepackage{longtable}
\usepackage{array}
\usepackage{xcolor}
\usepackage{colortbl}
\usepackage{tabularx}
\usepackage{booktabs}
\usepackage{tabularx}
\usepackage{array}
\usepackage[table]{xcolor}
\usepackage{ragged2e}
\usepackage[utf8]{inputenc}
\usepackage{amsmath, amssymb, amsfonts}
\usepackage{graphicx}
\usepackage{booktabs}
\newcommand{\lightrule}{\arrayrulecolor{gray!45}\specialrule{0.35pt}{0pt}{0pt}\arrayrulecolor{black}}
\usepackage{bm} 
\usepackage{rotating} 
\newcolumntype{Y}{>{\RaggedRight\arraybackslash}X}
\usepackage{ragged2e} 
\usepackage{dblfloatfix} 
\newcolumntype{Y}{>{\RaggedRight\arraybackslash}X}
\usepackage{algorithm}
\usepackage{algpseudocode}
\usepackage{etoolbox}
\AtBeginEnvironment{algorithm}{\footnotesize}
\algrenewcommand\algorithmicindent{1.0em}
\algrenewcommand\algorithmiccomment[1]{\hfill{\scriptsize$\triangleright$~#1}}
\newcommand{\tatt}{T_{\mathrm{att}}}

\algrenewcommand\algorithmicindent{0.9em}
\algrenewcommand\algorithmiccomment[1]{\hfill{\footnotesize$\triangleright$~#1}}
\newcommand{\Tatt}{T_{\mathrm{att}}} 
\title{Feature-Aware Anisotropic Local Differential Privacy for Utility-Preserving Graph Representation Learning in Metal Additive Manufacturing}

\begin{document}
\author{MD Shafikul Islam
    \affiliation{Graduate Research Assistant \\
	  Mechanical and Industrial Engineering\\
  Louisiana State University\\
  Baton Rouge, LA 70803 \\
  \texttt{misla79@lsu.edu}
    }	
}

\author{Mahathir Mohammad Bappy\thanks{} \\
    \affiliation{ Assistant Professor \\
    Mechanical and Industrial Engineering\\
  Louisiana State University\\
  Baton Rouge, LA 70803 \\
  \texttt{mmbappy@lsu.edu}
    }
}

\author{Saifur Rahman Tushar
    \affiliation{Graduate Research Assistant \\
	Mechanical and Industrial Engineering\\
  Louisiana State University\\
  Baton Rouge, LA 70803 \\
  \texttt{SaifurRahman.Tushar@lsu.edu}
    }	
}

\author{Md Arifuzzaman
    \affiliation{Assistant Professor\\
	Department of Computer Science \\
  Missouri University of Science and Technology\\
  Missouri S\&T, Rolla, MO 65409 \\
  \texttt{marifuzzaman@mst.edu}
    }	
}

\maketitle

\begin{abstract}
Metal additive manufacturing (AM) enables the fabrication of safety-critical components, but reliable quality assurance depends on high-fidelity sensor streams containing proprietary process information, limiting collaborative data sharing. Existing defect-detection models typically treat melt-pool observations as independent samples, ignoring layer-wise physical couplings, including heat accumulation and track interactions, that govern porosity formation. Moreover, conventional privacy-preserving techniques, particularly Local Differential Privacy (LDP), cause severe utility degradation due to uniform noise injection across all feature dimensions. To address these interrelated challenges, we propose \textbf{FI-LDP-HGAT}, a computational framework that combines two methodological components: a stratified Hierarchical Graph Attention Network (HGAT) that captures spatial and thermal dependencies across scan tracks and deposited layers, and a feature-importance-aware anisotropic Gaussian mechanism (FI-LDP) for non-interactive feature privatization. Unlike isotropic LDP, FI-LDP redistributes the privacy budget across embedding coordinates using an encoder-derived importance prior, assigning lower noise to task-critical thermal signatures and higher noise to redundant dimensions while maintaining formal $(\epsilon, \delta)$-LDP guarantees. Experiments on a Directed Energy Deposition (DED) porosity dataset demonstrate that FI-LDP-HGAT achieves 81.5\% utility recovery at a moderate privacy budget ($\epsilon = 4$) and maintains defect recall of 0.762 under strict privacy ($\epsilon = 2$), while outperforming classical ML, standard GNNs, and alternative privacy mechanisms including DP-SGD across all evaluated metrics. Mechanistic analysis confirms a strong negative correlation (Spearman $\rho = -0.81$) between feature importance and noise magnitude, providing interpretable evidence that the privacy--utility gains are driven by principled anisotropic allocation.
\end{abstract}


\begin{figure*}[t]
    \centering
    \includegraphics[width=1\linewidth]{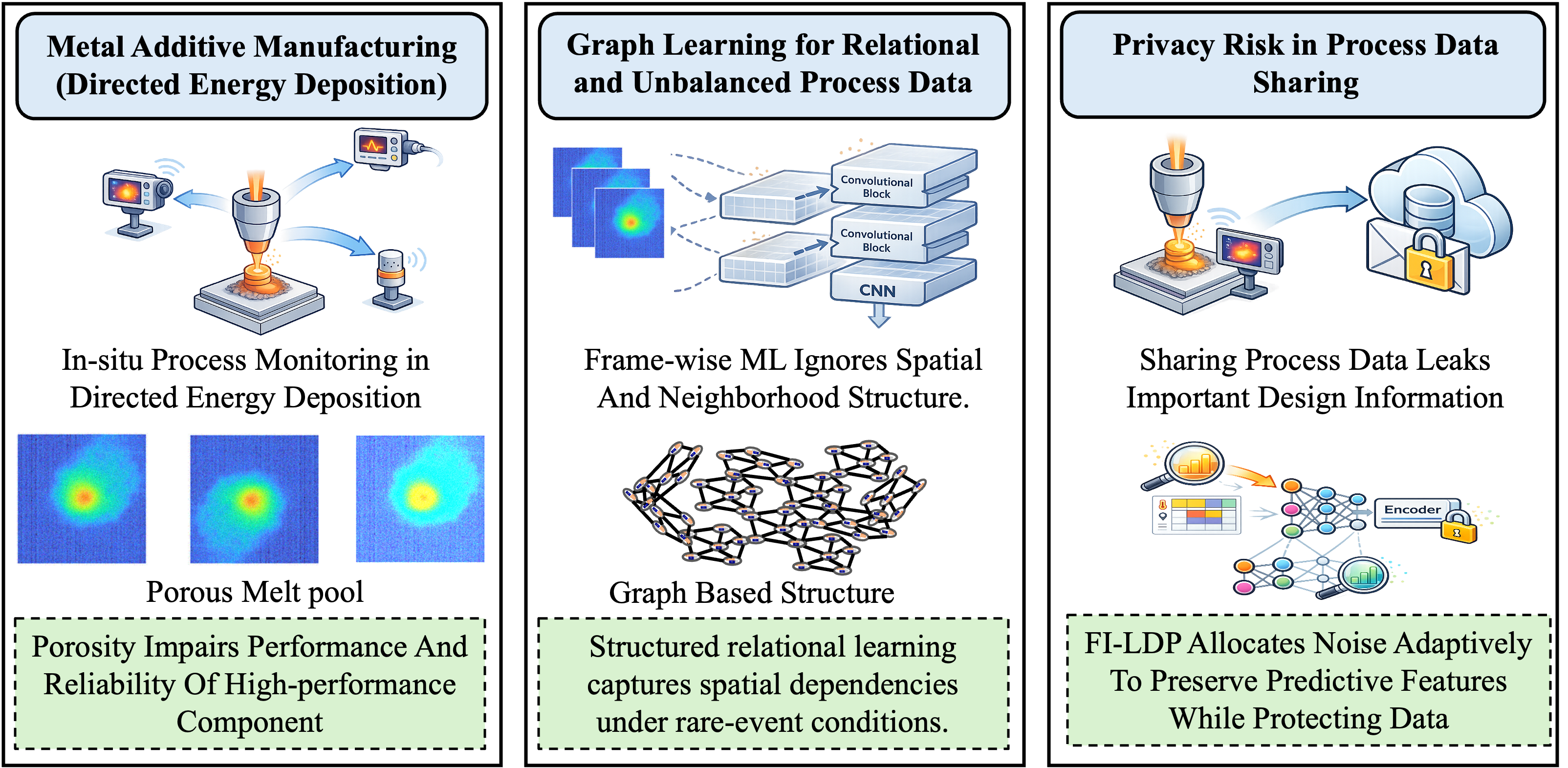}
    \vspace{-2mm}
    \caption{Framework for utility-preserving private feature release. Structured multimodal records are encoded, privatized via importance-aware anisotropic LDP (FI-LDP), and processed through a stratified Hierarchical Graph Attention Network (HGAT) for defect prediction.}
    \label{fig:overview_framework}
\end{figure*}

\section{Introduction}
\label{sec:intro}

Data-driven quality assurance in metal additive manufacturing (AM) increasingly depends on computational methods that can simultaneously model complex process physics and satisfy real-world deployment constraints such as data confidentiality~\cite{tian2021deep,chen2024situ}. In Directed Energy Deposition (DED) platforms, including Laser Engineered Net Shaping (LENS\texttrademark{}), layer-wise fabrication produces high-fidelity thermal and spatial sensor streams that encode information about melt-pool dynamics, heat accumulation, and defect propensity~\cite{zhang2019process,bappy2022morphological}. Among process-induced defects, \emph{porosity} remains one of the most persistent failure modes, degrading fatigue life and mechanical integrity and constituting a primary barrier to certification-grade deployment in aerospace and biomedical systems~\cite{ren2023machine,ansari2022convolutional,esfahani2022situ}.

A central computational limitation of existing learning pipelines is the \emph{independent-sample assumption}. Convolutional neural networks (CNNs), recurrent architectures (LSTMs), and classical machine learning models typically treat each melt-pool observation as an individual sample~\cite{ho2021dlam,estalaki2022predicting,mao2023deep}. This modeling choice disregards the physical coupling inherent to layer-wise AM: defect propensity at a given location is influenced by cumulative heat accumulation from adjacent scan tracks and the thermal history of underlying layers~\cite{haribaskar2024defects,rottler2025effect,ozel2023review}. Structured representations that explicitly encode these relational dependencies are needed to advance predictive capability beyond what frame-level models can achieve.

Graph Neural Networks (GNNs) offer a principled abstraction for such relational data, aggregating neighborhood information through learned message-passing operators to enable context-aware inference~\cite{wu2020comprehensive,wasi2024graph,velivckovic2017graph}. Graph-based inductive biases have shown promise in manufacturing settings when geometry- or process-aware priors are incorporated~\cite{mozaffar2021geometry,xiong2024knowledge,zhou2025spatially}. However, deploying graph learning in collaborative manufacturing ecosystems introduces a second challenge: sharing sensor-derived representations across organizations can expose proprietary ``process fingerprints'', thermal signatures, scan-path geometry, and design-specific parameter sets---that constitute a manufacturer's core competitive advantage~\cite{yang2020big,bappy2024toward,bappy2023privacy}. This tension between relational modeling and IP protection has motivated a growing body of work on privacy-preserving computational methods for manufacturing~\cite{bappy2025adaptive,lee2024privacy,shi2024sensor,oskolkov2025incremental,rahman2024taxonomy,ali2024layered}. Among formal approaches, \emph{local} differential privacy (LDP) is particularly attractive for decentralized settings because each data holder randomizes its own features before any downstream sharing~\cite{wang2020comprehensive,duchi2013local}. Yet standard LDP mechanisms rely on \emph{isotropic} perturbations that uniformly corrupt all coordinates, degrading task-critical signals and redundant dimensions alike in manufacturing embeddings where predictive utility is concentrated in a sparse subset of features~\cite{murakami2019utility,acharya2020context,huang2024enhancing,li2024survey}.

To address these interrelated computational challenges, this paper proposes \textbf{FI-LDP-HGAT}, a methodology that combines two computational components tailored to privacy-preserving graph learning in manufacturing: (i) \emph{Feature-Importance-guided Local Differential Privacy} (FI-LDP), an anisotropic Gaussian mechanism for non-interactive feature privatization, and (ii) a \emph{stratified Hierarchical Graph Attention Network} (HGAT) that encodes manufacturing-specific physical priors for structure-aware inference. FI-LDP redistributes privacy perturbation across feature dimensions using encoder-derived importance signals, assigning lower noise variance to task-critical coordinates and higher noise variance to redundant dimensions under a formal $(\epsilon,\delta)$-LDP accounting framework. The stratified HGAT constructs a layer-restricted hybrid $k$NN graph that couples in-layer spatial proximity with learned thermal embedding similarity, enabling attention-based message passing that respects the physical structure of the deposition process. Figure~\ref{fig:overview_framework} summarizes the motivating problem context that gives rise to FI-LDP-HGAT: structured defect prediction in DED requires relational learning, while collaborative data sharing requires formal privacy protection. The contributions of this work are three-fold:

\begin{enumerate}
    \item \textbf{Privacy mechanism design}: We develop FI-LDP, an importance-aware anisotropic Gaussian mechanism for local feature privatization. FI-LDP redistributes per-dimension privacy budgets using a temperature-controlled power-law allocation derived from a supervised warmup signal under a formal $(\epsilon,\delta)$-LDP accounting framework (Eq.~\eqref{eq:budget_allocation}). This distinguishes FI-LDP from both isotropic LDP~\cite{wang2020comprehensive} and heuristic de-identification approaches~\cite{bappy2025adaptive} by providing a principled, tunable mechanism that explicitly couples noise allocation to task utility.

    \item \textbf{Physics-informed computational modeling}: We design a layer-stratified hybrid graph construction and hierarchical attention architecture that encodes domain-specific manufacturing priors, intra-layer thermal coupling, spatial--thermal hybrid proximity, and edge-affinity-biased attention---into the computational model. Unlike standard GAT applied to generic graphs, this formulation restricts message passing to physically meaningful neighborhoods and integrates process-aware edge priors into the attention mechanism.

    \item \textbf{Comprehensive quantitative evaluation}: We evaluate the proposed framework on an experimental DED porosity dataset against baseline methods spanning classical machine learning, deep learning, graph learning, and privacy-preserving approaches. The results show that FI-LDP-HGAT maintains strong detection utility under source-side privacy constraints, achieving 81.5\% utility recovery relative to the non-private oracle at $\epsilon=4$ while preserving high rare-defect recall under stricter privacy budgets.
\end{enumerate}

The remainder of the paper is organized as follows. Section~\ref{sec:background} reviews related work on graph learning for AM, privacy-preserving computational methods in manufacturing, and local privacy mechanisms. Section~\ref{sec:method} presents the proposed framework. Section \ref{sec:setup} explains  experimental setup and data acquisition,  section~\ref{sec:results} reports experiments and privacy--utility analysis. Section~\ref{sec:discussion} discusses implications and future directions. Section~\ref{sec:conclusion} concludes.


\section{Background and Related Work}
\label{sec:background}

This section reviews the computational methods relevant to the three challenges that FI-LDP-HGAT is designed to address: (i) how existing porosity predictors model or fail to model the relational structure of AM process data; (ii) how graph learning captures that structure but introduces IP exposure risks in collaborative settings; and (iii) how current privacy-preserving methods for manufacturing fall short of formal, utility-aware feature privatization. The section concludes by identifying the specific methodological gap that the proposed framework targets.

\subsection{Learning-Based Porosity Prediction from In-situ Sensing}
\label{subsec:bg_porosity}

Data-driven porosity detection has progressed through several modeling paradigms. CNN-based architectures first demonstrated that melt-pool geometry carries discriminative signatures for defect classification from coaxial or infrared imagery~\cite{zhang2019process,ansari2022convolutional}. Temporal extensions such as CNN--LSTM architectures were subsequently introduced to capture dynamic thermal fluctuations across sequential frames~\cite{ho2021dlam,mao2023deep}, and multimodal fusion approaches improved defect assessment by combining multiple sensor streams~\cite{karthikeyan2023situ}. Classical machine learning methods have also established competitive baselines: Random Forests applied to engineered thermal descriptors for voxel-level prediction~\cite{estalaki2022predicting}, and Self-Organizing Maps (SOMs) for unsupervised melt-pool clustering that achieved up to 96\% detection accuracy on DED thin-wall builds~\cite{khanzadeh2019situ}. Stochastic defect localization using Gaussian mixture representations has further begun to address spatial correlation in cooperative AM settings~\cite{rescsanski2024stochastic}.

Despite these advances, the methods above share a common computational limitation: each melt-pool observation is modeled as an independent sample. This assumption prevents the model from exploiting track-to-track interactions and cumulative heat-accumulation effects, physical couplings that are central drivers of defect formation in DED-style deposition~\cite{ozel2023review,haribaskar2024defects,rottler2025effect}. Overcoming this limitation requires structured representations that explicitly encode spatial and layer-wise dependencies, which motivates graph-based formulations.

\begin{table*}[t]
\centering
\footnotesize
\renewcommand{\arraystretch}{1.25}
\setlength{\tabcolsep}{5pt}
\caption{Comparison of privacy-preserving methods relevant to manufacturing analytics. The table highlights the protection target, privacy mechanism, and the main limitation of each method relative to graph-ready feature release.}
\label{tab:privacy_comparison}

\begin{tabularx}{\textwidth}{@{}p{3.2cm}p{1.7cm}p{1.5cm}p{1.0cm}X@{}}
\toprule
\textbf{Method} & \textbf{Target} & \textbf{Privacy type} & \textbf{Formal} & \textbf{Main relevance / limitation} \\
\midrule

SIA+ASIG~\cite{bappy2025adaptive}
& Raw images
& Heuristic
& No
& De-identifies melt-pool images through stochastic augmentation and surrogate generation, but does not provide a formal privacy bound for learned embeddings. \\
\lightrule

MNP~\cite{lee2024privacy}
& Model weights
& $(\epsilon,\delta)$-DP
& Yes
& Perturbs model parameters during distributed training; protects the model rather than released feature representations. \\
\lightrule

Blockchain/Encryption~\cite{shi2024sensor,oskolkov2025incremental}
& Data in transit
& Access control
& No
& Ensures integrity and secure transmission, but does not address statistical privacy or utility-aware feature perturbation. \\
\lightrule

Federated learning~\cite{wang2026privacy,zhou2025privacy}
& Training data
& Varies
& Optional
& Avoids raw-data centralization, but requires iterative communication and is not designed for single-shot feature release. \\
\lightrule

\textbf{FI-LDP (Proposed)}
& Feature embeddings
& $(\epsilon,\delta)$-LDP
& Yes
& Applies importance-guided anisotropic noise to graph-ready feature embeddings, enabling formal privacy with downstream graph learning utility. \\
\bottomrule
\end{tabularx}
\end{table*}
\subsection{Graph Representation Learning for Structured Manufacturing Data}
\label{subsec:bg_graph}

Graph representation learning addresses the independent-sample limitation by representing sensor observations as nodes and physically meaningful relations: spatial proximity, layer adjacency, or thermal similarity as edges. GNNs leverage iterative neighborhood aggregation to propagate context across connected nodes, while attention-based variants (GATs) learn data-adaptive aggregation weights that prioritize informative neighbors under varying thermal regimes~\cite{wu2020comprehensive,velivckovic2017graph}. In the AM domain, Mozaffar et al.~\cite{mozaffar2021geometry} developed a geometry-agnostic GNN for thermal modeling along DED scan paths, demonstrating that graph inductive biases improve generalization across part geometries. Zhou et al.~\cite{zhou2025spatially} proposed a spatially-informed GNN with multiphysics priors for online surface deformation prediction in digital twinning applications. Graph-theoretic frameworks have also been applied to manufacturing cybersecurity risk modeling, illustrating the broader applicability of graph-based computational methods in manufacturing systems~\cite{rahman2024taxonomy}.

These models, however, uniformly assume access to high-fidelity, unperturbed features. In cross-organization collaboration, raw features or learned embeddings may encode proprietary process information, creating a fundamental tension between the relational modeling capability of GNNs and the data confidentiality requirements of multi-stakeholder manufacturing. Resolving this tension requires privacy mechanisms that can protect released features without destroying the embedding geometry on which graph construction and attention depend.

\begin{figure*}[t]
    \centering
    \includegraphics[width=\linewidth]{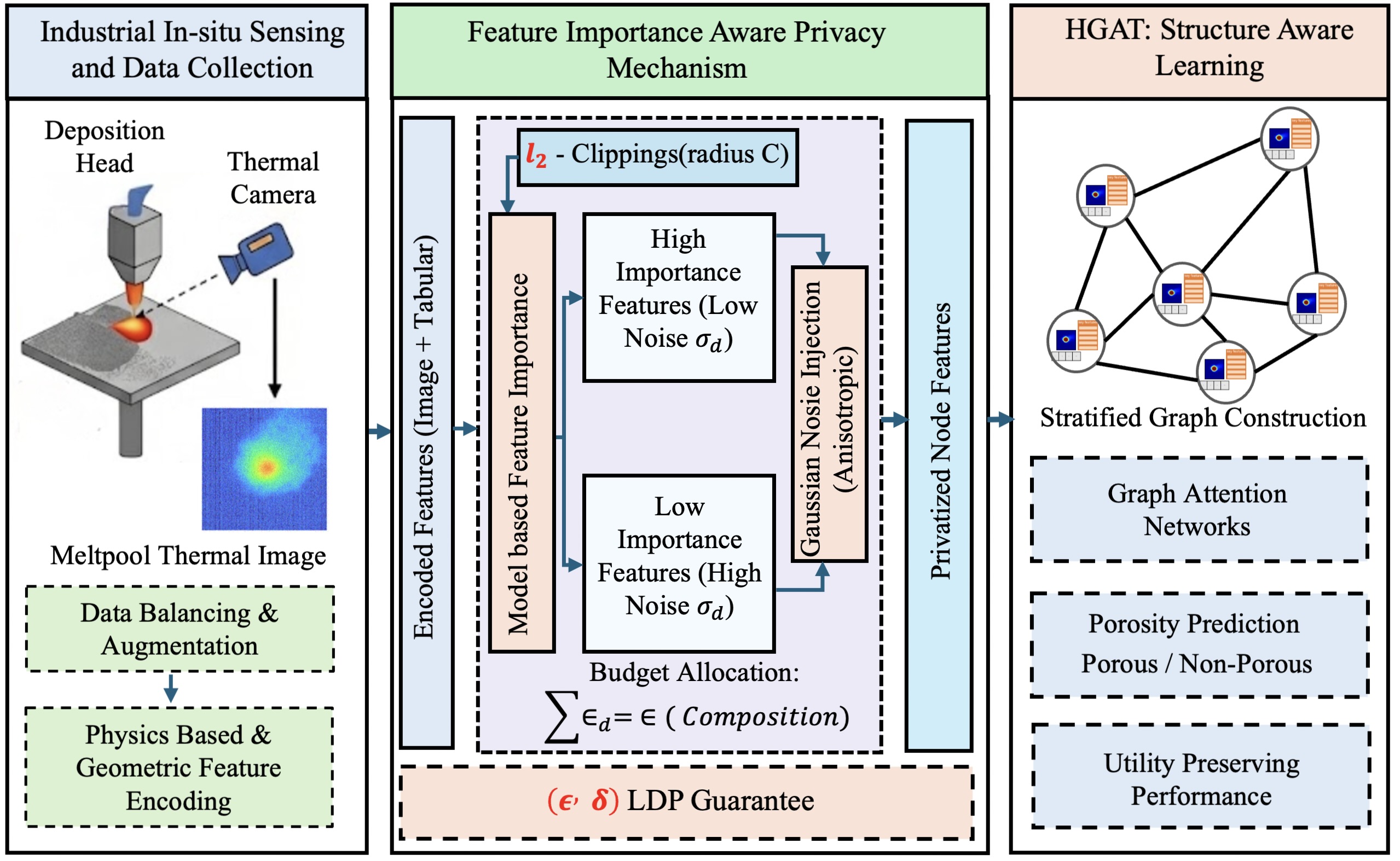}
    \caption{High-level overview of the utility-preserving private graph learning framework for in-situ porosity prediction. The pipeline integrates (i) porous-targeted data augmentation to address class imbalance, (ii) a supervised warmup phase for feature importance estimation, and (iii) importance-aware anisotropic Local Differential Privacy (FI-LDP) for secure feature release to a stratified Hierarchical Graph Attention Network (HGAT).}
    \label{fig:methodology_flow}
\end{figure*}

\subsection{Privacy-Preserving Computational Methods in Manufacturing Analytics}
\label{subsec:bg_privacy_mfg}

The need to balance collaborative data sharing with IP protection has driven the development of several privacy-preserving approaches for manufacturing, which can be organized by their protection target (Table~\ref{tab:privacy_comparison}). At the \emph{image level}, Bappy et al.~\cite{bappy2025adaptive} proposed an adaptive de-identification method for DED thermal data that combines stochastic image augmentation with surrogate image generation to mask printing trajectory information while preserving defect-modeling utility. This approach operates directly on raw melt-pool images and provides empirical privacy, but it does not offer formal guarantees, and the utility--privacy trade-off depends on an augmentation policy rather than on a provable bound. At the \emph{model level}, Lee et al.~\cite{lee2024privacy} introduced Mosaic Neuron Perturbation (MNP), which perturbs neural network parameters during distributed training to prevent model inversion attacks under differential privacy. MNP protects the model rather than the data, making it complementary to feature-release mechanisms but inapplicable when encoded features must be shared for downstream graph construction. At the \emph{infrastructure level}, blockchain-based frameworks have been proposed for securing sensor data in transit~\cite{shi2024sensor,oskolkov2025incremental}; these ensure data integrity and access control but do not address the statistical utility--privacy trade-off inherent to feature perturbation. Finally, \emph{federated learning} approaches for AM enable collaborative model training without centralizing raw data~\cite{wang2026privacy,zhou2025privacy}, but they require iterative multi-round communication and do not support the non-interactive, single-shot feature-release setting considered in this work. Beyond model- and data-level privacy mechanisms, prior work has emphasized that additive manufacturing information requires protection strategies that go beyond conventional encryption, especially when sensitive process knowledge may still be exposed through side-channel or workflow-level leakage~\cite{lubell2022protecting}. This broader AM security perspective reinforces the need for formal, utility-aware feature privatization mechanisms for collaborative analytics. But none of these methods directly address the problem of releasing \emph{learned, graph-ready feature embeddings} under formal local privacy guarantees while preserving task-relevant structure for downstream attention-based inference. This is the specific computational gap that FI-LDP is designed to fill.

\begin{table*}[t]
\centering
\footnotesize
\renewcommand{\arraystretch}{1.15}
\setlength{\tabcolsep}{6pt}
\caption{\textbf{Nomenclature.} Key symbols used in the proposed FI-LDP-HGAT framework.}
\label{tab:nomenclature}
\begin{tabularx}{\textwidth}{@{}p{3.1cm}X@{}}
\toprule
\textbf{Symbol} & \textbf{Description} \\
\midrule
$\mathcal{G}=(\mathcal{V},\mathcal{E})$ & Layer-stratified process graph with nodes $\mathcal{V}$ and edges $\mathcal{E}$. \\
$v_i$ & Node corresponding to a localized melt-pool observation. \\
$\xi_i=(I_i,\mathbf{s}_i,\mathbf{g}_i,y_i)$ & Multimodal record: thermal patch, process-state features, geometric context, and label. \\
$I_i\in\mathbb{R}^{H\times W}$ & In-situ thermal image patch (melt-pool neighborhood). \\
$\mathbf{s}_i\in\mathbb{R}^{d_s}$ & Process-state / melt-pool scalar descriptors. \\
$\mathbf{g}_i\in\mathbb{R}^{d_g}$ & Geometric context (layer index and in-layer coordinates; optional part/toolpath attributes). \\
$y_i\in\{0,1\}$ & Porosity label after XCT-to-in-situ registration. \\
$\mathbf{z}^{(\mathrm{img})}_i=f_{\theta}(I_i)$ & Image embedding (thermal fingerprints) from ResNet-18 encoder. \\
$\mathbf{z}^{(\mathrm{ctx})}_i=g_{\phi}(\mathbf{s}_i,\mathbf{g}_i)$ & Context embedding from MLP over process and geometry features. \\
$\mathbf{x}_i=[\mathbf{z}^{(\mathrm{img})}_i\|\mathbf{z}^{(\mathrm{ctx})}_i]\in\mathbb{R}^{D}$ & Fused node feature used for warmup, privatization, and graph learning. \\
$\ell_i$ & Physical layer index of node $v_i$ (enforces within-layer edges). \\
$D_{ij}$ & Hybrid distance for $k$NN edges (spatial proximity + thermal embedding similarity). \\
$k,\alpha,\tau$ & Graph hyperparameters: neighbors, mixing, kernel bandwidth. \\
$\Tatt$ & Attention temperature used in HGAT (softmax/logit scaling). \\
$w_{ij}$ & Edge affinity prior, $w_{ij}=\exp(-D_{ij}/\tau)$. \\
$\mathbf{q}\in\mathbb{R}^{D}$ & Global feature-importance prior from warmup head weights. \\
$(\epsilon,\delta)$ & Local differential privacy parameters for feature release. \\
$C_{\mathrm{img}},C_{\mathrm{ctx}},C_{\mathrm{tot}}$ & Modality clipping bounds and fused sensitivity bound. \\
$\epsilon_d,\delta_d,\sigma_d$ & Dimension-wise privacy budget and FI-LDP Gaussian noise scale. \\
$\hat{\mathbf{x}}_i$ & Privatized feature vector released under FI-LDP. \\
$a^{(g)}_{ij}$ & HGAT attention coefficient at graph layer $g$ for neighbor aggregation. \\
$t^*$ & Validation-tuned decision threshold for node-level classification. \\
\bottomrule
\end{tabularx}
\end{table*}

\begin{figure*}[t]
    \centering
    \includegraphics[width=\linewidth]{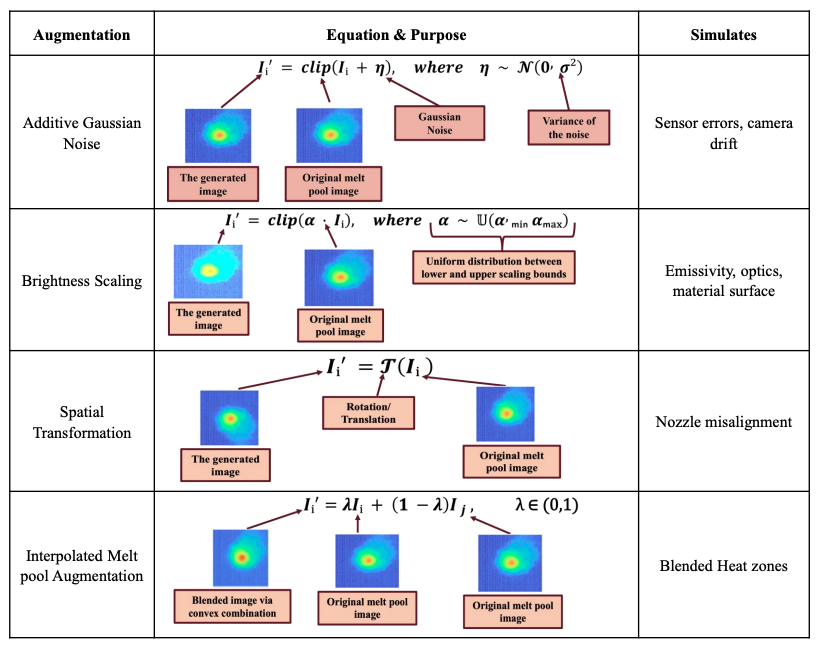}
    \caption{\textbf{Porous-targeted augmentation operators used during training.}
    The four transformations (A1)--(A4) are applied \emph{only} to minority-class (porous) thermal frames to increase defect-signal diversity under severe class imbalance. The operators emulate sensor noise, intensity drift, mild registration errors, and interpolated defect signatures; validation and test splits remain unaugmented.}
    \label{fig:data_augmentation}
\end{figure*}

\subsection{Local Differential Privacy for Continuous Feature Release}
\label{subsec:bg_ldp}

Local differential privacy (LDP) requires each data holder to randomize its own record before release, removing the need for a trusted curator~\cite{wang2020comprehensive,duchi2013local}. For continuous features, the standard Gaussian mechanism adds isotropic noise calibrated to the $\ell_2$-sensitivity of the released vector~\cite{zheng2024overview}. While this provides a clean formal guarantee, isotropic perturbation treats all feature coordinates uniformly---a mismatch with manufacturing embeddings where a small number of dimensions (e.g., peak melt-pool temperature, eccentricity) carry most of the predictive signal.

In the broader machine learning community, several works have explored privacy-preserving graph neural networks. Sajadmanesh and Gatica-Perez~\cite{sajadmanesh2021locally} proposed locally private GNNs with node-level LDP, and subsequent work introduced aggregation perturbation mechanisms for differentially private graph learning~\cite{sajadmanesh2023gap}. These methods target social-network-style graphs with discrete or low-dimensional attributes and do not address the high-dimensional multimodal embeddings, severe class imbalance, or manufacturing-specific graph topologies encountered in AM process monitoring. Utility-aware LDP mechanisms that allocate noise according to feature contribution have been explored in distribution estimation settings~\cite{murakami2019utility,acharya2020context}, but their integration with graph learning and manufacturing-domain priors remains unexplored. FI-LDP addresses this gap by deriving a per-dimension noise schedule from a supervised warmup signal, providing a principled bridge between feature importance and privacy budget allocation that is compatible with downstream graph construction and attention-based inference.

\subsection{Research Gaps and Positioning}
\label{subsec:bg_gaps}

Table~\ref{tab:privacy_comparison} contrasts the proposed FI-LDP with existing privacy-preserving approaches across several computational dimensions. Taken together, the literature reveals a methodological gap at the intersection of structured relational inference and source-side local privacy. Graph-based predictors are well suited to capture layer-wise and spatial coupling in AM process streams, but they generally assume non-private access to high-fidelity features. Standard local differential privacy, by contrast, provides formal protection but remains utility-agnostic, uniformly perturbing the embedding geometry that graph construction and attention mechanisms depend on. Existing privacy-preserving methods in manufacturing further focus on image-level de-identification, model-level perturbation, or infrastructure-level security, none of which directly address formal, non-interactive privatization of graph-ready feature representations. FI-LDP-HGAT is designed to bridge this gap by combining a stratified graph model with an importance-guided anisotropic LDP mechanism for utility-preserving feature release under formal $(\epsilon,\delta)$-LDP guarantees.

\begin{figure*}[t]
    \centering
    \includegraphics[width=\linewidth]{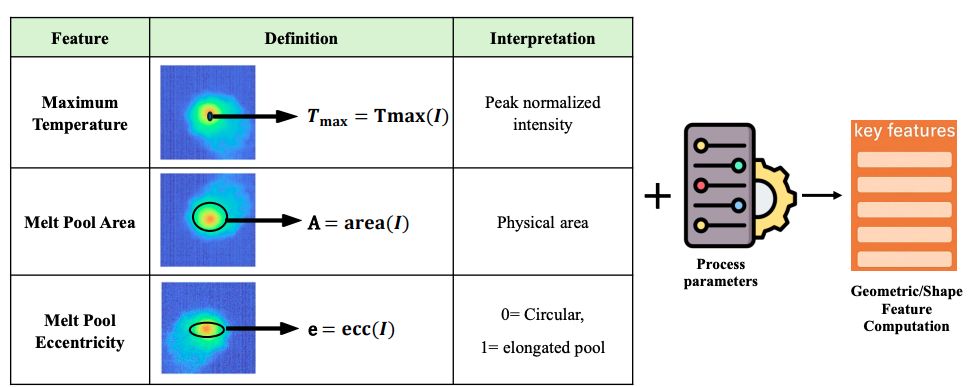}
    \vspace{-2mm}
    \caption{\textbf{Multimodal feature extraction and fusion for node representation.}
    Each node feature $\mathbf{x}_i$ concatenates an image-derived embedding $\mathbf{z}^{(\mathrm{img})}_i=f_{\theta}(I_i)$ (thermal fingerprints) with a context embedding $\mathbf{z}^{(\mathrm{ctx})}_i=g_{\phi}(\mathbf{s}_i,\mathbf{g}_i)$ capturing process-state and geometric descriptors. This fused representation is used for warmup importance estimation and as the input to FI-LDP privatization in subsequent stages.}
    \label{fig:feature_fusion}
\end{figure*}

\begin{figure*}[b]
    \centering
    \includegraphics[width=\linewidth]{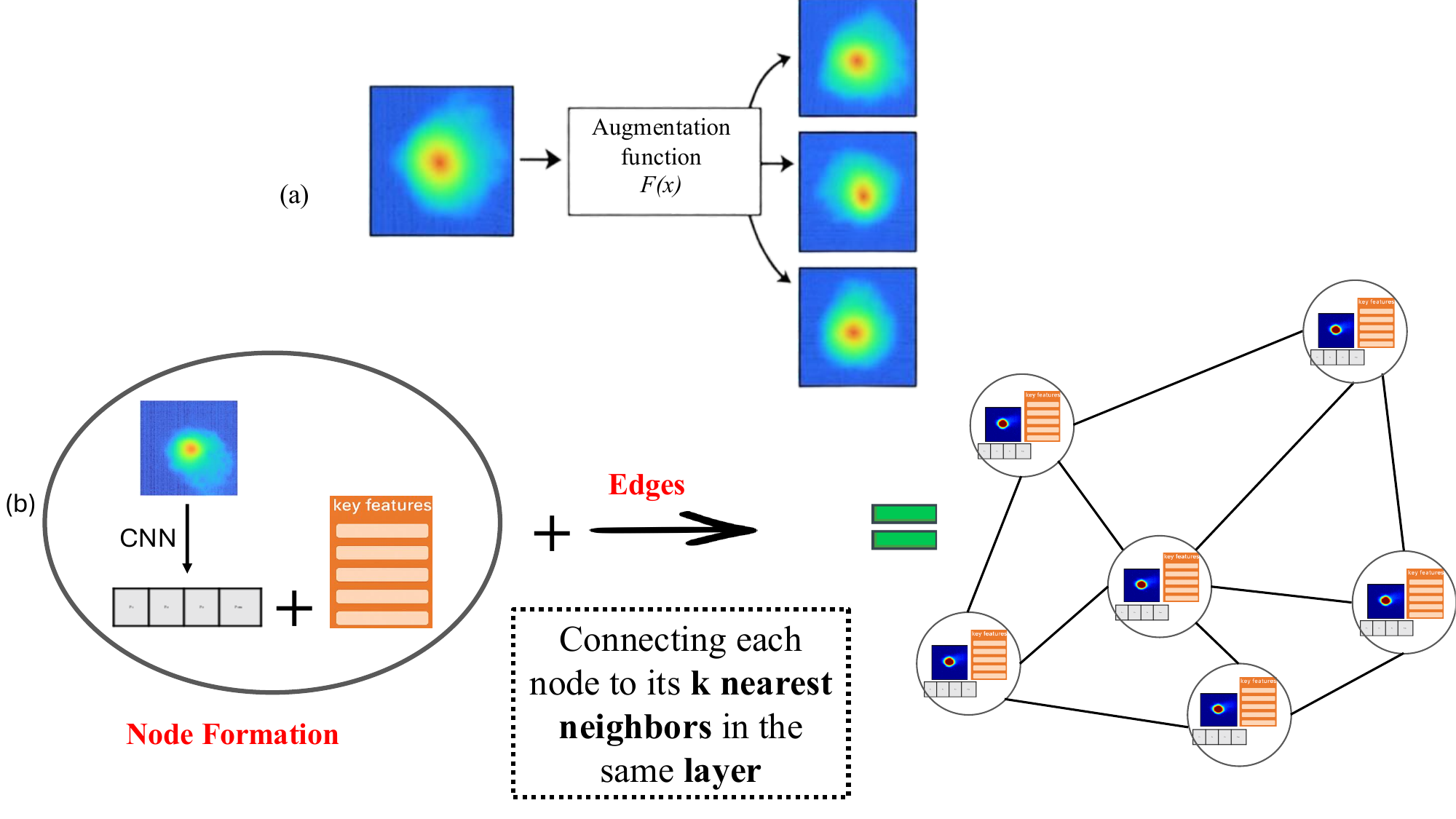}
    \caption{\textbf{Graph construction pipeline for FI-LDP-HGAT.}
    (a) Porous-targeted augmentation applies transformation operators $F(\cdot)$ to minority-class thermal patches to increase defect-signal diversity during training (Sec.~\ref{subsec:imbalance}).
    (b) Node and edge formation for the layer-stratified hybrid $k$NN graph: each node aggregates an image embedding from the thermal patch and a context vector of process/geometric features (Sec.~\ref{subsec:extraction}). 
    Within each physical layer, edges connect each node to its $k$ nearest neighbors under the hybrid distance in Eq.~\eqref{eq:hybrid_distance}, combining in-layer spatial proximity and thermal embedding similarity.}
    \label{fig:graph_construction}
\end{figure*}

\section{Methodology}
\label{sec:method}

We develop a utility-preserving private analytics pipeline for in-situ porosity prediction. The framework follows a staged design that (i) increases defect-signal diversity under extreme class imbalance via porous-targeted augmentation, (ii) learns multimodal representations and estimates a global importance prior for importance-weighted privatization, and (iii) enables structure-aware inference on a layer-stratified process graph. The overall workflow is summarized in Fig.~\ref{fig:methodology_flow}.

\subsection{Graph Formulation for Node-Level Porosity Inference}
\label{subsec:problem}

We formulate in-situ porosity detection in layer-wise metal AM as a node-level binary classification problem on a stratified process graph $\mathcal{G}=(\mathcal{V},\mathcal{E})$. Each node $v_i\in\mathcal{V}$ corresponds to a localized melt-pool observation
\begin{equation}
\xi_i := \big(I_i,\ \mathbf{s}_i,\ \mathbf{g}_i,\ y_i\big),
\end{equation}
where $I_i \in \mathbb{R}^{H\times W}$ is an in-situ thermal image patch centered at the deposition zone. The vector $\mathbf{s}_i\in\mathbb{R}^{d_s}$ collects scalar process-state and melt-pool descriptors (e.g., intensity/area statistics and sensing-derived summaries). The vector $\mathbf{g}_i\in\mathbb{R}^{d_g}$ encodes geometric and spatial context, including the physical layer index and in-layer coordinates (and, when available, part/toolpath-related attributes). The label $y_i\in\{0,1\}$ indicates whether porosity is present at the corresponding location, obtained by registering post-process XCT pore annotations to the in-situ observation within a fixed spatial tolerance. We define a fused node feature vector as
\begin{equation}
\mathbf{x}_i=\big[\mathbf{z}^{(\mathrm{img})}_i \parallel \mathbf{z}^{(\mathrm{ctx})}_i\big]\in\mathbb{R}^{D},
\end{equation}
where $\mathbf{z}^{(\mathrm{img})}_i$ is a learned embedding extracted from $I_i$ (thermal fingerprints), and $\mathbf{z}^{(\mathrm{ctx})}_i$ is an embedding of process and geometric context derived from $(\mathbf{s}_i,\mathbf{g}_i)$.

\subsection{Porous-Targeted Augmentation for Rare-Event Sensing}
\label{subsec:imbalance}

Porosity formation in metal AM is sparse and spatially localized. As a result, naive empirical risk minimization tends to learn a majority-dominated boundary and under-detect rare defects. To increase sensitivity to porous regions without distorting the nominal (non-porous) distribution, we adopt a porous-targeted augmentation protocol that operates only on minority-class observations during training.

\textit{Targeted augmentations.} For each porous thermal frame $I_i$, we apply one of the following on-the-fly transformations (Fig.~\ref{fig:data_augmentation}) to emulate realistic sensing noise and modest process drift. Let $\clip(\cdot)$ denote intensity clipping to the valid sensor range \cite{gonzalez2009digital}.

\textbf{(A1) Additive Gaussian noise.} This models camera stochasticity and acquisition drift so the encoder learns defect morphology that is stable under pixel-level perturbations \cite{shorten2019survey}:
\begin{equation}
\begin{aligned}
I_i' &= \clip\!\big(I_i+\boldsymbol{\eta}\big),\\
\boldsymbol{\eta} &\sim \Normal(\mathbf{0},\sigma^2\mathbf{I}),\quad
\sigma^2 \sim \Unif[\sigma_{\min}^2,\sigma_{\max}^2].
\end{aligned}
\end{equation}
Here $\boldsymbol{\eta}$ is i.i.d.\ pixel noise and $\sigma^2$ is sampled per augmentation instance to span a plausible range of sensor noise levels.

\textbf{(A2) Brightness scaling.} This captures global intensity shifts (e.g., emissivity/exposure variation) to reduce reliance on absolute thermal magnitude \cite{shorten2019survey, chua2017process}:
\begin{equation}
\begin{aligned}
I_i' &= \clip(a\,I_i),\\
a &\sim \Unif[a_{\min},a_{\max}].
\end{aligned}
\end{equation}
The random scalar $a$ enforces invariance to multiplicative intensity changes while preserving melt-pool shape cues.

\textbf{(A3) Rotation/translation.} This approximates mild registration drift relative to the nominal toolpath, so predictions are robust to small pose/alignment errors:
\begin{equation}
\begin{aligned}
I_i' &= \mathcal{S}_{\Delta u,\Delta v}\!\big(\mathcal{R}_{\theta}(I_i)\big),\\
\theta &\sim \Unif[-\theta_{\max},\theta_{\max}],\\
\Delta u,\Delta v &\sim \Unif[-\Delta_{\max},\Delta_{\max}].
\end{aligned}
\end{equation}
$\mathcal{R}_{\theta}$ applies a bounded in-plane rotation and $\mathcal{S}_{\Delta u,\Delta v}$ applies a bounded translation; together they mimic small coordinate misalignment without changing the underlying defect structure.

\textbf{(A4) Interpolated melt-pool synthesis.} This densifies the porous manifold by generating intermediate defect signatures via convex mixing \cite{zhang2017mixup}:
\begin{equation}
\begin{aligned}
I_i' &= \clip\!\big(\lambda I_i+(1-\lambda)I_j\big),\\
\mathbf{s}_i' &= \lambda \mathbf{s}_i+(1-\lambda)\mathbf{s}_j,\\
\lambda &\sim \Unif[\lambda_{\min},\lambda_{\max}].
\end{aligned}
\end{equation}
Here, $i$ and $j$ are sampled from the porous set to avoid synthesizing majority-class patterns. The mixing weight $\lambda$ samples points within the convex hull of observed porous instances, yielding plausible intermediate patterns in image space and corresponding process-state descriptors. All augmentations are applied only to the training split; validation and test data remain unaugmented to ensure unbiased evaluation.

\subsection{In-situ Feature Extraction and Warmup for Importance-Weighted Privatization}
\label{subsec:extraction}

The porous-targeted augmentation is designed to ensure that rare defect signatures contribute sufficient gradient signal during training. We next leverage this strengthened minority signal to obtain a stable \emph{importance prior} for FI-LDP. Concretely, we (i) learn multimodal embeddings that capture thermal patterns and geometry-aware process context, and (ii) run a short supervised warmup stage to estimate which coordinates of the fused representation are most predictive of porosity. This warmup stage is performed \emph{before} local privatization and is used only to calibrate the privacy mechanism.

\paragraph{Morphological encoding of in-situ thermal signatures.}
A ResNet-18 backbone $f_{\theta}(\cdot)$ maps each in-situ thermal frame $I_i$ to a compact embedding \cite{he2016deep}:
\begin{equation}
\mathbf{z}^{(\mathrm{img})}_i=f_{\theta}(I_i)\in\mathbb{R}^{d_{\mathrm{img}}}.
\end{equation}
The intent is to encode melt-pool footprint patterns associated with porosity.

\paragraph{Context-aware encoding of process and geometric features.}
We encode the process-state descriptors $\mathbf{s}_i$ and geometric features $\mathbf{g}_i$ via an MLP after standardization (and optional low-order interaction features):
\begin{equation}
\mathbf{z}^{(\mathrm{ctx})}_i=g_{\phi}(\mathbf{s}_i,\mathbf{g}_i)\in\mathbb{R}^{d_{\mathrm{ctx}}}.
\end{equation}
This pathway captures layer-wise spatial context and local energy/stability cues, which influence defect likelihood through heat accumulation and track-to-track interactions. We then form the multimodal node feature used downstream by concatenation:
\begin{equation}
\mathbf{x}_i=\big[\mathbf{z}^{(\mathrm{img})}_i \parallel \mathbf{z}^{(\mathrm{ctx})}_i\big]\in\mathbb{R}^{D}.
\end{equation}

\begin{figure*}[t]
\centering
\includegraphics[width=\linewidth]{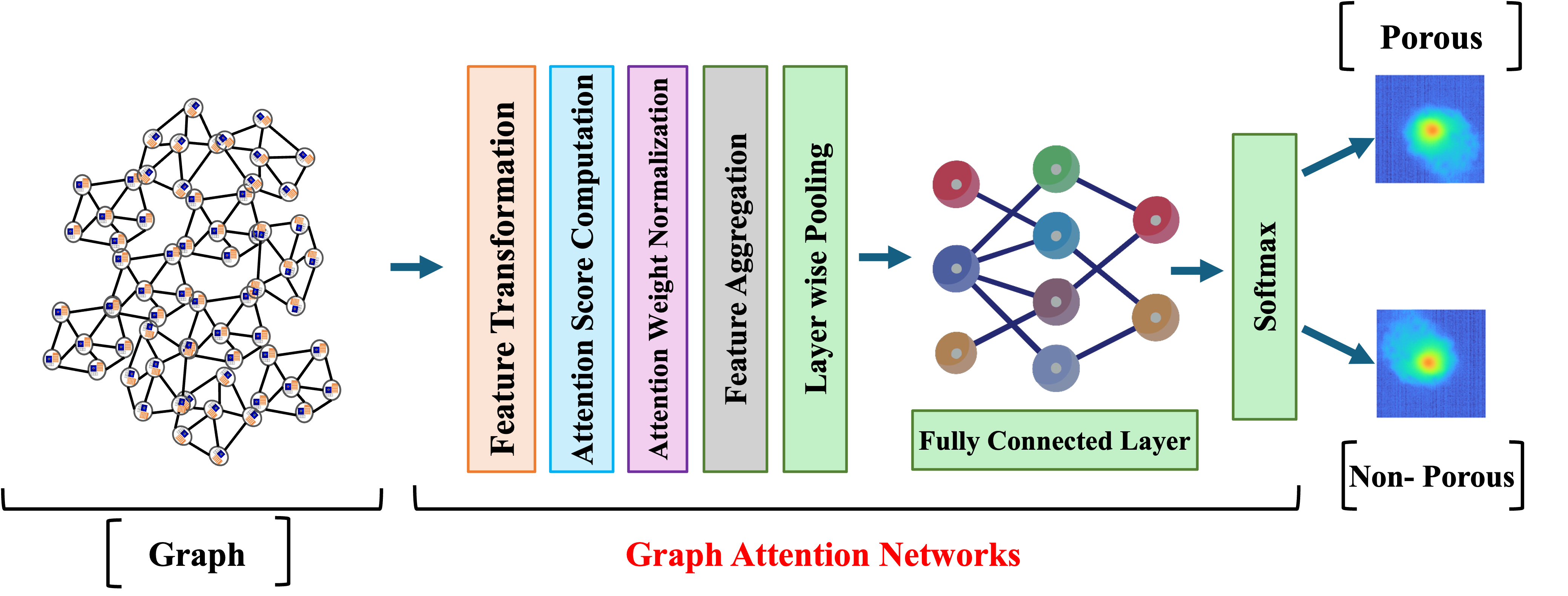}
\caption{\textbf{HGAT message passing with attention.}
 Edge priors bias attention toward spatially/thermally consistent neighbors, while learned coefficients adaptively weight neighborhood contributions for node-level porosity inference.}
\label{fig:gat_overview}
\end{figure*}

Using the same imbalance-aware sampling policy as Sec.~\ref{subsec:imbalance}, we train a lightweight classifier head on $\mathbf{x}_i$ for a short warmup phase (with label smoothing) to stabilize porosity-discriminative directions in representation space \cite{szegedy2016rethinking}. The warmup head is a linear classifier trained with cross-entropy (label smoothing) for $E_{\mathrm{warm}}$ epochs under the same imbalance-aware sampling policy. Let $\mathbf{W}\in\mathbb{R}^{H\times D}$ denote the warmup projection weights. We compute a global importance vector $\mathbf{q}\in\mathbb{R}^{D}$ as
\begin{equation}
q_d=\frac{1}{H}\sum_{h=1}^{H}\big|W_{hd}\big|,\qquad d=1,\dots,D,
\end{equation}
where larger $q_d$ indicates coordinates consistently exploited to separate porous from non-porous observations under the rare-event training regime. The vector $\mathbf{q}$ is aggregated over the warmup training set and is used only to allocate privacy noise in FI-LDP; it is not released at the record level.

\subsection{Layer-Stratified Hybrid Graph Construction and Hierarchical Graph Attention Learning}
\label{subsec:graph}

Thermal transport and melt-pool interactions in layer-wise metal AM induce strong \emph{intra-layer} coupling: neighboring tracks within the same layer share a similar heat-accumulation history and often exhibit correlated defect propensity. We encode this manufacturing prior by constructing a \emph{layer-stratified} process graph and performing node-level inference via hierarchical attention-based message passing. Figure~\ref{fig:graph_construction} illustrates the node/edge formation workflow.

\paragraph{Layer-stratified hybrid $k$NN graph construction.}
We define a stratified graph $\mathcal{G}=(\mathcal{V},\mathcal{E})$ by restricting edges to within-layer neighborhoods,
\begin{equation}
(i,j)\in\mathcal{E}\ \Rightarrow\ \ell_i=\ell_j,
\label{eq:within_layer_edges}
\end{equation}
where $\ell_i$ is the physical layer index. Within each layer $\ell$, we connect each node $i$ to its $k$ nearest neighbors under a hybrid distance that combines (i) in-layer spatial proximity and (ii) similarity of learned thermal embeddings \cite{wang2019dynamic}.

\noindent\textit{Cosine similarity.}
\begin{equation}
\cos(\mathbf{u},\mathbf{v})
:=\frac{\mathbf{u}^\top\mathbf{v}}{\|\mathbf{u}\|_2\,\|\mathbf{v}\|_2}.
\label{eq:cosine_def}
\end{equation}

Let $\mathbf{p}_i^{yz}=(y_i,z_i)$ denote the in-layer coordinates of node $i$. The hybrid distance between nodes $i$ and $j$ is
\begin{equation}
D_{ij}
=\alpha\left\|\mathbf{p}_i^{yz}-\mathbf{p}_j^{yz}\right\|_2
+(1-\alpha)\Big(1-\cos(\mathbf{z}^{(\mathrm{img})}_i,\mathbf{z}^{(\mathrm{img})}_j)\Big),
\label{eq:hybrid_distance}
\end{equation}
where $\alpha\in[0,1]$ controls the geometry--appearance trade-off \cite{wang2019dynamic}. We convert $D_{ij}$ into a soft edge-affinity prior via a heat kernel,
\begin{equation}
w_{ij}=\exp\!\left(-\frac{D_{ij}}{\tau}\right),
\label{eq:edge_prior}
\end{equation}
where $\tau>0$ is the neighborhood bandwidth. The prior $w_{ij}$ biases learning toward nearby and thermally similar events while retaining flexibility through attention.

\paragraph{Hierarchical graph attention (HGAT) for structure-aware inference.}
Let $\mathbf{h}_i^{(g)}$ denote the node representation at HGAT layer $g$ (distinct from $\ell_i$). For neighbor $j\in\mathcal{N}(i)$, we compute attention logits by combining transformed features with the edge prior \cite{velivckovic2017graph}:
\begin{equation}
e_{ij}^{(g)}=
\mathrm{LReLU}\!\left(
\mathbf{a}^\top
\big[
\mathbf{W}\mathbf{h}_i^{(g)} \parallel
\mathbf{W}\mathbf{h}_j^{(g)} \parallel
w_{ij}
\big]
\right),
\end{equation}
followed by normalized coefficients
\begin{equation}
a_{ij}^{(g)}=\frac{\exp(e_{ij}^{(g)})}{\sum_{r\in\mathcal{N}(i)}\exp(e_{ir}^{(g)})},
\end{equation}
and aggregation
\begin{equation}
\mathbf{h}_i^{(g+1)}=
\sigma\!\left(
\sum_{j\in\mathcal{N}(i)} a_{ij}^{(g)}\,\mathbf{W}\mathbf{h}_j^{(g)}
\right),
\end{equation}
where $\sigma(\cdot)$ is a nonlinearity. In practice, we use multi-head attention and concatenate (or average) heads at each layer (Fig.~\ref{fig:gat_overview}). The final node score $\hat{p}_i\in(0,1)$ is obtained via a classifier head. Training uses a weighted focal loss to emphasize rare porous nodes, and the operating threshold is tuned on validation ($t^*$) to maximize F1 \cite{lin2017focal}.

\begin{figure*}[t]
    \centering
    \includegraphics[width=\linewidth]{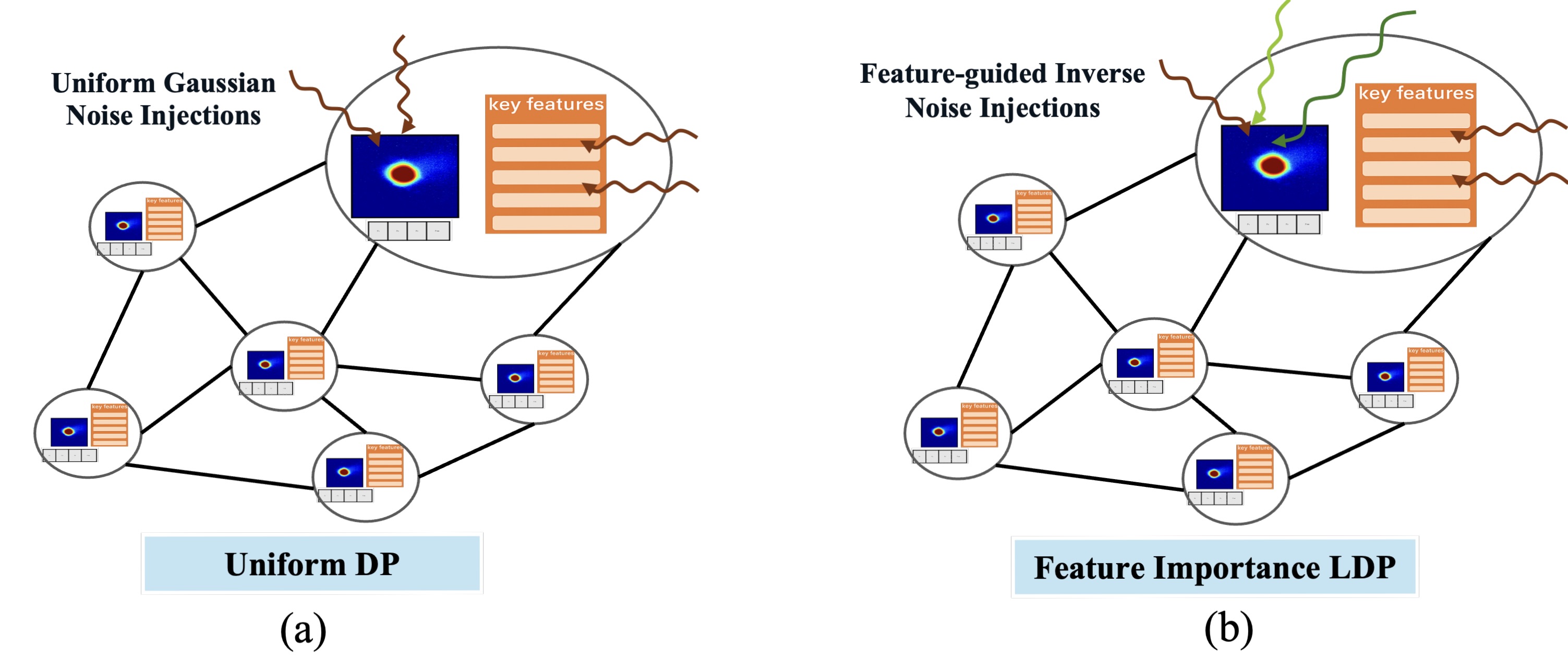}
    \caption{\textbf{Effect of isotropic vs.\ importance-guided local privatization on graph features.}
    (a) \textit{Uniform-LDP (isotropic Gaussian noise).} A single privacy budget is enforced by injecting the same noise scale into every feature coordinate, perturbing task-critical ``thermal fingerprint'' dimensions and redundant dimensions equally. This uniform corruption distorts the relative geometry of node embeddings and can weaken attention-based neighborhood aggregation.
    (b) \textit{FI-LDP (importance-guided anisotropic noise).} The privacy budget is redistributed across coordinates using the warmup-derived importance prior $\mathbf{q}$, assigning smaller noise to high-importance coordinates and larger noise to low-importance coordinates under the same $(\epsilon,\delta)$ guarantee. This preserves task-relevant subspaces while maintaining local privacy for feature release.}
    \label{fig:fi_ldp_mechanism}
\end{figure*}
 
\subsection{FI-LDP: Importance-Weighted Local Feature Privatization}
\label{subsec:fi_ldp}

To support collaborative analytics without releasing raw process fingerprints, we enforce privacy at the \emph{feature-release} boundary \cite{duchi2013local}. After extracting the fused representation $\mathbf{x}_i\in\mathbb{R}^{D}$, each facility releases only a privatized vector $\hat{\mathbf{x}}_i$ and performs all subsequent steps (graph construction and HGAT training/inference) on $\hat{\mathbf{x}}_i$. This yields a \emph{non-interactive} protocol (single-shot release). Figure~\ref{fig:fi_ldp_mechanism} summarizes the FI-LDP mechanism.

\paragraph{Modality-aware clipping (bounded sensitivity).}
We first bound record-level sensitivity by $\ell_2$-clipping each modality embedding:
\begin{equation}
\begin{aligned}
\tilde{\mathbf{x}}_i
&=\Big[
\clip\big(\mathbf{z}^{(\mathrm{img})}_i,C_{\mathrm{img}}\big)\ \Big\|\ 
\clip\big(\mathbf{z}^{(\mathrm{ctx})}_i,C_{\mathrm{ctx}}\big)
\Big],\\
\clip(\mathbf{u},C)
&=\frac{\mathbf{u}}{\max\!\left(1,\ \|\mathbf{u}\|_2/C\right)} .
\end{aligned}
\label{eq:clipping}
\end{equation}
This implies a fused bound
\begin{equation}
C_{\mathrm{tot}}=\sqrt{C_{\mathrm{img}}^2+C_{\mathrm{ctx}}^2},
\label{eq:ctot}
\end{equation}
which is used to calibrate the noise scale.

\begin{algorithm}
\caption{FI-LDP-HGAT: Utility-Preserving Private Porosity Prediction}
\label{alg:framework}
\footnotesize
\begin{algorithmic}[1]
\Require Records $\{\xi_i=(I_i,\mathbf{s}_i,\mathbf{g}_i,y_i)\}_{i=1}^{N}$; privacy $(\epsilon,\delta)$;
FI-LDP $(\beta,\eta,C_{\mathrm{img}},C_{\mathrm{ctx}})$; graph $(k,\alpha,\tau)$
\Ensure Predictions $\{\hat{y}_i\}_{i=1}^{N}$

\State \textbf{Stage 0: Warmup to estimate importance prior.}
\State Train encoders $(f_{\theta},g_{\phi})$ using porous-targeted augmentation
\State Fit a warmup linear head with weights $\mathbf{W}\in\mathbb{R}^{H\times D}$
\State $q_d \gets \frac{1}{H}\sum_{h=1}^{H}\lvert W_{hd}\rvert,\quad d=1,\ldots,D$ \Comment{global importance prior}

\vspace{1mm}
\State \textbf{Stage 1: Pre-compute FI-LDP noise scales.}
\State $C_{\mathrm{tot}}\gets \sqrt{C_{\mathrm{img}}^{2}+C_{\mathrm{ctx}}^{2}}$
\State $Z\gets \sum_{r=1}^{D}(q_r+\eta)^{\beta}$
\For{$d=1$ to $D$}
    \State $\epsilon_d \gets \epsilon\,\frac{(q_d+\eta)^{\beta}}{Z}$;\quad $\delta_d \gets \delta/D$
    \State $\sigma_d \gets \frac{2C_{\mathrm{tot}}\sqrt{2\ln(1.25/\delta_d)}}{\epsilon_d}$
\EndFor
\State $\boldsymbol{\sigma}\gets(\sigma_1,\ldots,\sigma_D)$

\vspace{1mm}
\State \textbf{Stage 2: FI-LDP feature release (non-interactive).}
\For{$i=1$ to $N$}
    \State $\mathbf{z}^{(\mathrm{img})}_i \gets f_{\theta}(I_i)$;\ \ $\mathbf{z}^{(\mathrm{ctx})}_i \gets g_{\phi}(\mathbf{s}_i,\mathbf{g}_i)$
    \State $\tilde{\mathbf{x}}_i \gets \big[\clip(\mathbf{z}^{(\mathrm{img})}_i,C_{\mathrm{img}})\ \|\ \clip(\mathbf{z}^{(\mathrm{ctx})}_i,C_{\mathrm{ctx}})\big]$
    \State $\hat{\mathbf{x}}_i \gets \tilde{\mathbf{x}}_i + \boldsymbol{\nu}_i,\ \ \boldsymbol{\nu}_i\sim \Normal(\mathbf{0},\mathrm{diag}(\boldsymbol{\sigma}^2))$
\EndFor

\vspace{1mm}
\State \textbf{Stage 3: Stratified graph construction and HGAT inference.}
\State For each physical layer $\ell$, build a $k$NN graph using hybrid distance $D_{ij}$ and set $w_{ij}=\exp(-D_{ij}/\tau)$
\State Train HGAT on $\mathcal{G}=(\mathcal{V},\mathcal{E})$ using weighted focal loss; tune threshold $t^*$ on validation
\For{$i=1$ to $N$}
    \State $\hat{y}_i \gets \mathbb{I}\{\mathrm{HGAT}(\hat{\mathbf{x}}_i)>t^*\}$
\EndFor
\end{algorithmic}
\end{algorithm}

\paragraph{Importance-weighted privacy budget allocation.}
Uniform-LDP perturbs all coordinates equally, which is inefficient when predictive utility is concentrated in a small subset of dimensions. Using the warmup-derived importance prior $\mathbf{q}\in\mathbb{R}^{D}$ (Sec.~\ref{subsec:extraction}), FI-LDP allocates a larger share of the privacy budget to high-importance coordinates. Let $\beta\ge 0$ control anisotropy and $\eta>0$ stabilize allocation:
\begin{equation}
\begin{aligned}
Z &= \sum_{r=1}^{D}(q_r+\eta)^{\beta},\\
\epsilon_d &= \epsilon\,\frac{(q_d+\eta)^{\beta}}{Z},\qquad
\delta_d = \delta/D,\qquad d=1,\dots,D .
\end{aligned}
\label{eq:budget_allocation}
\end{equation}
Thus, larger $q_d$ yields larger $\epsilon_d$ and hence weaker perturbation on task-critical coordinates.

\paragraph{Anisotropic Gaussian perturbation (local release).}
We privatize each record by adding coordinate-wise Gaussian noise:
\begin{equation}
\begin{aligned}
\hat{\mathbf{x}}_i &= \tilde{\mathbf{x}}_i+\boldsymbol{\nu}_i,\qquad
\nu_{i,d}\sim \Normal(0,\sigma_d^2),\\
\sigma_d &= \frac{2C_{\mathrm{tot}}\sqrt{2\ln\!\big(1.25/\delta_d\big)}}{\epsilon_d}.
\end{aligned}
\label{eq:anisotropic_gaussian}
\end{equation}
Equations~\eqref{eq:budget_allocation}--\eqref{eq:anisotropic_gaussian} make explicit that FI-LDP is equivalent to injecting an \emph{anisotropic} noise vector with dimension-dependent scales $\{\sigma_d\}$ (Fig.~\ref{fig:fi_ldp_mechanism}). After feature release, all downstream processing is purely a function of $\{\hat{\mathbf{x}}_i\}$, and therefore does not incur additional privacy loss by post-processing immunity. For consistency with the deployment setting, we apply the same privatization mechanism to train/validation/test representations under the reported $(\epsilon,\delta)$ budgets.

%
%

%

\section{Experimental Setup and Data Acquisition}
\label{sec:setup}

The proposed framework is evaluated using an experimental dataset of Ti-6Al-4V thin-walled structures (dimensions: $25.4 \times 1.0 \times 12.7$~mm) fabricated via an OPTOMEC LENS\texttrademark{} 750 Directed Energy Deposition (DED) system. The specimens were built using a laser power of 400~W and a constant travel speed of 10.58~mm/s \cite{zamiela2023thermal}. During the build, a Stratonics dual-wavelength pyrometer was integrated into the system to provide a top-down, on-axis view of the deposition zone. The sensor captured high-resolution thermal images ($200 \times 200$ pixels) at a sampling rate of 6.7~fps, focusing on the melt pool and the adjacent Heat-Affected Zone (HAZ). The pyrometer was calibrated for a range of 1000--2500$^\circ$C, ensuring the capture of critical thermal gradients during the solidification phase. Post-fabrication, internal porosity was characterized via X-ray Computed Tomography (XCT). To maintain high fidelity, pores ranging from 0.05~mm to 1.00~mm in diameter were cataloged. 

To establish ground truth labels, the in-process thermal signatures were registered with the XCT-detected pore locations. We applied a spatial tolerance of 0.5 mm to accommodate thermal expansion during build-up and slight coordinate system misalignments between the pyrometer and XCT reference frames. The final dataset comprises 1,564 unique observations. In alignment with high-quality stable manufacturing, the data exhibits a severe class imbalance: only 70 frames are labeled as porous ($4.47\%$), while 1,494 instances are non-porous.



\section{Results}
\label{sec:results}

We evaluate FI-LDP-HGAT along five axes: (i) non-private baseline benchmarking to establish the performance landscape, (ii) privacy-aware benchmarking against alternative protection mechanisms, (iii) mechanism-level evidence that FI-LDP allocates noise anisotropically as intended, (iv) privacy--utility scaling over a range of budgets, and (v) ablations that isolate the contribution of individual design choices. All results are computed on a fixed stratified split (60/20/20) and averaged across five random seeds unless otherwise noted.

\subsection{Non-Private Baseline Comparison}
\label{subsec:baselines}

Before evaluating privacy mechanisms, we first establish the non-private performance landscape by comparing the proposed HGAT architecture against representative baselines from classical machine learning, deep learning, and graph learning. All methods operate on the same dataset and train/val/test split. Table~\ref{tab:baseline_nonprivate} summarizes the results.

\begin{table}[h]
\centering
\caption{Non-private baseline comparison. All methods use the same data split. Feature columns indicate the input representation: \emph{Img} = ResNet-18 image embedding; \emph{Multi} = image + process-state + geometric context; \emph{Graph} = layer-stratified $k$NN topology.}
\label{tab:baseline_nonprivate}
\resizebox{\columnwidth}{!}{
\begin{tabular}{l c c c c c c}
\toprule
\textbf{Method} & \textbf{Features} & \textbf{AUC} & \textbf{AUPR} & \textbf{Rec@0.5} & \textbf{F1$^*$} & \textbf{F1$^*$ std} \\
\midrule
\multicolumn{7}{l}{\textit{Classical / deep learning (no graph structure)}} \\
SVM (RBF)           & Img   & 0.979 & 0.973 & 0.926 & 0.903 & 0.000 \\
MLP (2-layer)       & Img   & 0.986 & 0.948 & 0.778 & 0.824 & 0.022 \\
ResNet-18 + MLP     & Img   & 0.978 & 0.969 & 0.879 & 0.861 & 0.022 \\
\midrule
\multicolumn{7}{l}{\textit{Graph neural networks (with layer-stratified topology)}} \\
GCN                 & Multimodal + Graph & 0.964 & 0.931 & 0.946 & 0.862 & 0.022 \\
Vanilla GAT         & Multimodal + Graph & 0.977 & \textbf{0.967} & 0.960 & 0.854 & 0.030 \\
\textbf{HGAT (Ours)}& \textbf{Multimodal + Graph} & \textbf{0.990} & 0.907 & \textbf{N/A} & \textbf{0.941} & --- \\
\bottomrule
\end{tabular}}
\end{table}

Several observations are worth noting. First, flat classifiers (SVM, MLP, ResNet-18~+~MLP) operating on pre-extracted image embeddings achieve strong AUC values (0.978--0.986) and AUPR values (0.948--0.973). This reflects the discriminative power of the ResNet-18 encoder on thermal image patches, consistent with prior CNN-based porosity detection work~\cite{zhang2019process,ho2021dlam} rather than the contribution of the downstream classifier. However, these methods operate on image embeddings alone and do not incorporate process-state or geometric context, nor do they model relational dependencies across observations.

Second, among graph-based methods, GCN and vanilla GAT achieve high recall ($>$0.94) but lower tuned F1$^*$ (0.862 and 0.854, respectively), indicating that standard message-passing without edge priors or process-aware construction tends to over-predict the minority class. The proposed HGAT, which integrates multimodal features, layer-stratified hybrid edges, and edge-affinity-biased attention, achieves the highest calibrated F1$^*$ (0.941) among all methods, demonstrating that manufacturing-informed graph construction and attention design translate into measurable gains in defect discrimination. Critically, the non-private comparison is not the primary evaluation axis of this work. Flat classifiers require access to unperturbed, high-fidelity embeddings, a condition that is violated in multi-stakeholder manufacturing where IP-sensitive features must be privatized before release. The key question is how well each architecture degrades when privacy constraints are imposed, which we examine next.

\subsection{Privacy-Aware Benchmarking}
\label{subsec:privacy_bench}

We now evaluate FI-LDP-HGAT against alternative privacy mechanisms under matched $(\epsilon,\delta)$ budgets. Table~\ref{tab:privacy_benchmark} reports results at two representative privacy levels: a strict budget ($\epsilon=2.0$) and a moderate budget ($\epsilon=4.0$). We compare: (i) Uniform-LDP with HGAT (isotropic Gaussian perturbation on embeddings), (ii) DP-SGD-style training with HGAT (gradient-level clipping and Gaussian noise during optimization), and (iii) FI-LDP with HGAT (the proposed importance-guided anisotropic perturbation). The non-private HGAT oracle is included as an upper bound.

\begin{table}[h]
\centering
\caption{Privacy-aware method comparison at fixed privacy budgets. All methods use the same HGAT backbone and graph topology for fair comparison. $F1^*$ denotes the test F1-score at a validation-optimized threshold.}
\label{tab:privacy_benchmark}
\resizebox{\columnwidth}{!}{
\begin{tabular}{l c c c c c}
\toprule
\textbf{Method} & $\epsilon$ & \textbf{AUC} & \textbf{AUPR} & \textbf{Rec@0.5} & \textbf{F1$^*$} \\
\midrule
Non-Private Oracle       & ---  & 0.990 & 0.907 & N/A   & 0.941 \\
\midrule
DP-SGD + HGAT            & 2.0  & 0.436 & 0.130 & 0.489 & 0.194 \\
Uniform-LDP + HGAT       & 2.0  & 0.884 & 0.621 & 0.711 & 0.685 \\
\textbf{FI-LDP + HGAT (Ours)} & \textbf{2.0} & \textbf{0.913} & \textbf{0.664} & \textbf{0.762} & \textbf{0.686} \\
\midrule
DP-SGD + HGAT            & 4.0  & 0.446 & 0.132 & 0.489 & 0.202 \\
Uniform-LDP + HGAT       & 4.0  & 0.910 & 0.697 & 0.770 & 0.752 \\
\textbf{FI-LDP + HGAT (Ours)} & \textbf{4.0} & \textbf{0.936} & \textbf{0.751} & \textbf{0.777} & \textbf{0.767} \\
\bottomrule
\end{tabular}}
\end{table}

The results reveal a clear performance hierarchy under privacy constraints. DP-SGD-style training, which adds calibrated noise to gradients during optimization, suffers catastrophic utility loss (AUC~$\approx0.44$, F1$^*<0.21$) at both budget levels. This is consistent with known challenges of DP-SGD in high-dimensional, imbalanced regimes~\cite{huang2024enhancing}: gradient clipping combined with per-step noise injection disrupts the delicate optimization dynamics required for rare-event detection, causing the model to converge to near-majority-class prediction. This finding motivates the \emph{feature-release} privacy paradigm adopted by FI-LDP, where noise is injected once into the learned embedding rather than accumulated across training iterations. Uniform-LDP provides a substantially stronger baseline, achieving AUC~$=0.884$ and F1$^*=0.685$ at $\epsilon=2.0$. Under the same budget, FI-LDP improves AUC by 3.3\% (0.913 vs.\ 0.884), AUPR by 6.9\% (0.664 vs.\ 0.621), and recall by 7.2\% (0.762 vs.\ 0.711). At $\epsilon=4.0$, FI-LDP achieves F1$^*=0.767$, corresponding to 81.5\% utility recovery relative to the non-private oracle ($0.767/0.941$). These gains are consistent across metrics and support the hypothesis that importance-guided anisotropic perturbation preserves task-relevant subspaces more effectively than uniform corruption.


\begin{figure*}[b]
\centering
\includegraphics[width=\textwidth]{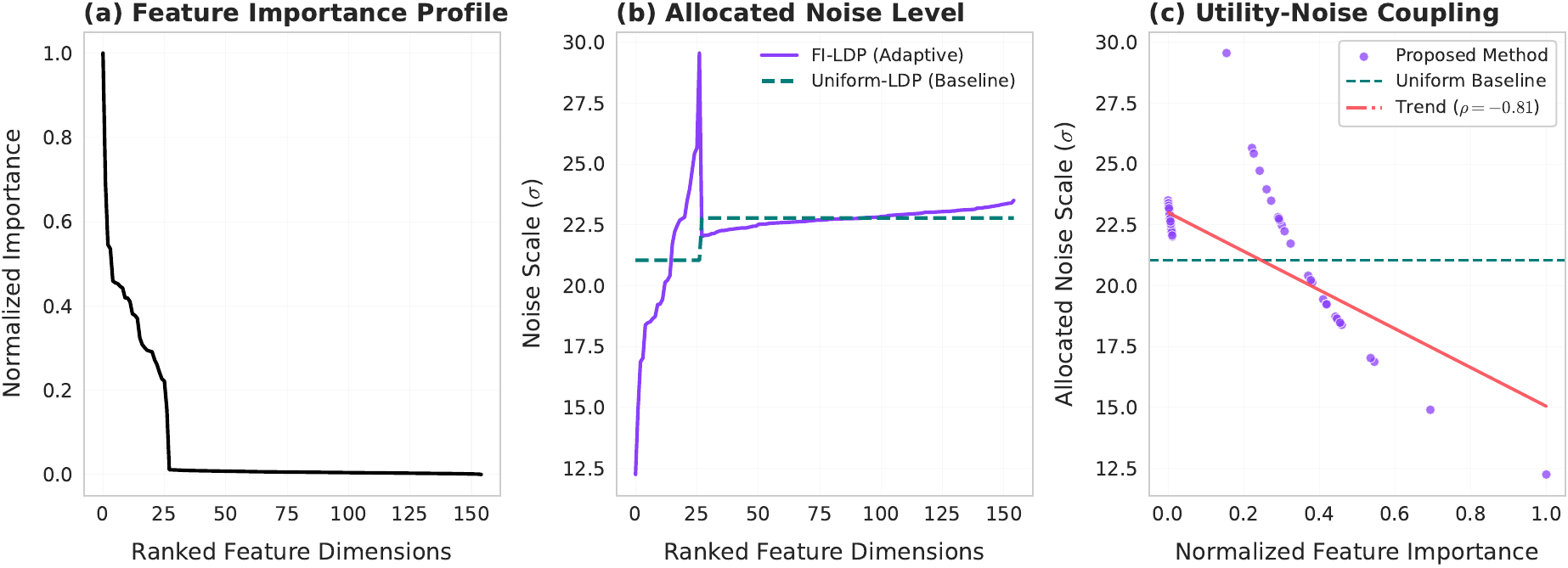}
\caption{\textbf{Mechanism insight for FI-LDP at $\epsilon=2.0$: importance-guided anisotropic perturbation.}
(a) Normalized importance scores (sorted) show that predictive utility is concentrated in a small subset of embedding coordinates.
(b) Allocated Gaussian noise scale $\sigma_j$ per coordinate: Uniform-LDP applies a constant $\sigma$, whereas FI-LDP assigns smaller $\sigma_j$ to high-importance coordinates and larger $\sigma_j$ to low-importance ones under the same privacy budget.
(c) Utility--noise coupling: scatter of importance vs.\ allocated $\sigma_j$ with a fitted trend, showing a strong negative monotonic association (Spearman $\rho=-0.81$), i.e., FI-LDP perturbs informative coordinates less aggressively.}
\label{fig:mechanism_insight}
\end{figure*}

\subsection{Mechanism Insight: Importance-Guided Anisotropic Noise Allocation}
To directly validate that FI-LDP implements importance-guided perturbation, we visualize how feature importance translates into dimension-wise noise allocation at $\epsilon=2.0$ in Fig.~\ref{fig:mechanism_insight}. The figure is constructed from the learned embedding coordinates, with dimensions sorted by descending importance. Fig.~\ref{fig:mechanism_insight}(a) shows a sharply heavy-tailed importance profile: importance drops rapidly within the first few ranked dimensions and then approaches a near-flat tail. This shape indicates that predictive utility is concentrated in a small subset of coordinates, while the majority of dimensions contribute marginal information. This concentration motivates anisotropic perturbation, since uniform corruption wastes the privacy budget on weak coordinates while unnecessarily degrading strong ones.

Fig.~\ref{fig:mechanism_insight}(b) reports the corresponding allocated noise scale $\sigma_j$ per ranked dimension. Uniform-LDP appears as a flat horizontal line, reflecting utility-blind isotropic perturbation with identical noise across all coordinates. In contrast, FI-LDP exhibits a structured, non-uniform profile: the noise scale decreases across the high-importance head and increases toward the low-importance tail. This redistribution provides mechanistic evidence that FI-LDP preserves task-critical coordinates by assigning them lower variance while pushing more perturbation into redundant dimensions under the same $(\epsilon,\delta)$ constraint. Fig.~\ref{fig:mechanism_insight}(c) further quantifies this coupling by plotting $\sigma_j$ against the (normalized) importance scores. The downward trend and the strong negative monotonic association (Spearman $\rho=-0.81$) confirm that FI-LDP systematically assigns less noise to more important features. Taken together, these provide interpretable evidence that the observed privacy--utility gains are driven by principled anisotropic allocation, not by incidental hyperparameter effects.

\begin{figure*}[t]
    \centering
    \includegraphics[width=\linewidth]{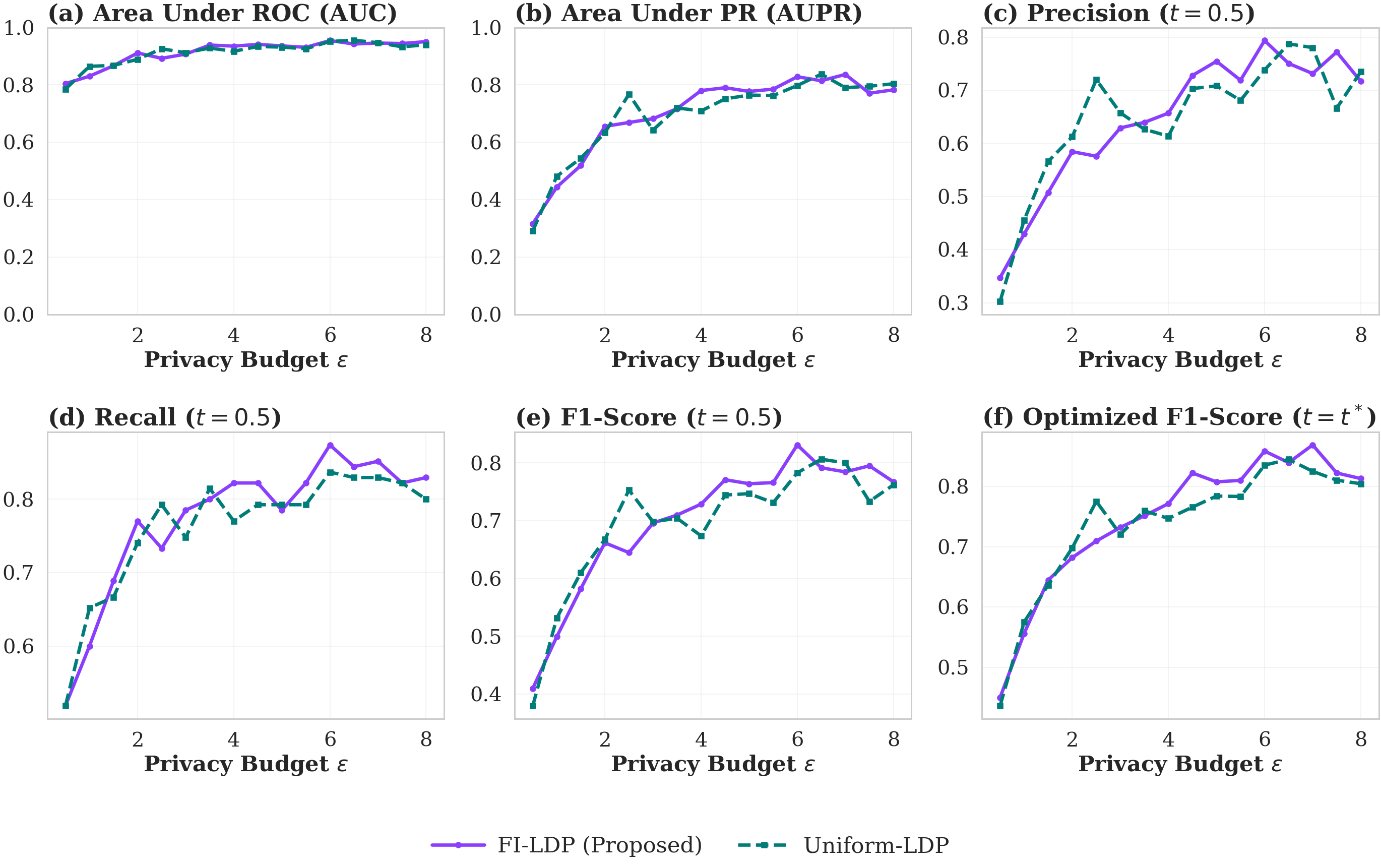}
    \caption{\textbf{Privacy--utility frontier for FI-LDP and Uniform-LDP.}
    Metrics are reported across $\epsilon \in [0.5, 8.0]$ under a fixed protocol (same split; seed-averaged runs). 
    (a) AUC and (b) AUPR summarize threshold-free ranking quality. 
    (c) Precision@0.5, (d) Recall@0.5, and (e) F1@0.5 are computed at the default decision threshold $t=0.5$. 
    (f) $F1^*$ is evaluated at a validation-optimised threshold $t^*$, representing the best calibrated operating point. 
    FI-LDP exhibits a higher utility ceiling in the moderate-privacy regime ($\epsilon\approx2$--$4$), while both methods converge at larger budgets as privacy-induced distortion diminishes.}
    \label{fig:privacy_metrics}
\end{figure*}

\subsection{Analysis of the Privacy--Utility Frontier}
We next sweep $\epsilon \in [0.5, 8.0]$ to characterize stability under varying privacy requirements. Fig.~\ref{fig:privacy_metrics} reports six complementary views of the privacy--utility frontier: (a) AUC and (b) AUPR summarize threshold-free ranking quality, (c) Precision@0.5 and (d) Recall@0.5 quantify the default operating point ($t=0.5$), (e) F1@0.5 summarizes the default precision--recall balance, and (f) the optimized $F1^*$ reports the best achievable operating point after tuning the threshold on validation ($t=t^*$). Table~\ref{tab:sweep} provides the corresponding FI-LDP values.

Across all metrics, FI-LDP remains informative even at strict budgets. At $\epsilon=1.0$, FI-LDP achieves AUC$=0.826$ and AUPR$=0.429$, indicating non-trivial ranking utility under strong noise. Utility increases as $\epsilon$ becomes more permissive (less noise), with diminishing returns beyond $\epsilon \ge 4$ (e.g., AUPR improves from $0.751$ at $\epsilon=4$ to $0.766$ at $\epsilon=8$). The separation between FI-LDP and Uniform-LDP is most visible in the moderate regime ($\epsilon\approx 2$--$4$), where FI-LDP attains higher AUPR (e.g., $0.664$ at $\epsilon=2$ and $0.751$ at $\epsilon=4$), reflecting improved preservation of rare-defect ranking signal. Small metric fluctuations across adjacent $\epsilon$ values are expected due to stochastic perturbations and finite-sample estimation on an imbalanced test set; the aggregate trend remains consistent, and the gap to Uniform-LDP narrows at larger budgets as both mechanisms inject less noise and approach the oracle~\cite{khavkin2025differential}.

\vspace{2mm}

\begin{table}[h]
\centering
\caption{FI-LDP utility scaling across privacy budgets. Recovery~\% is computed relative to $F1^*_{\text{oracle}}=0.941$.}
\label{tab:sweep}
\resizebox{\columnwidth}{!}{
\begin{tabular}{c c c c c c}
\toprule
$\epsilon$ & \textbf{AUC} & \textbf{AUPR} & \textbf{F1 ($t=0.5$)} & \textbf{F1$^*$ (tuned)} & \textbf{Recovery \%} \\
\midrule
0.5 & 0.792 & 0.302 & 0.361 & 0.440 & 46.7\% \\
1.0 & 0.826 & 0.429 & 0.490 & 0.537 & 57.0\% \\
2.0 & 0.913 & 0.664 & 0.647 & 0.685 & 72.8\% \\
4.0 & 0.936 & 0.751 & 0.677 & 0.767 & 81.5\% \\
8.0 & 0.943 & 0.766 & 0.776 & 0.801 & 85.1\% \\
\bottomrule
\end{tabular}}
\vspace{-4mm}
\end{table}

\subsection{Ablation Study and Sensitivity Analysis}
To attribute the observed privacy-utility gains to specific design choices, we conduct controlled ablations at a fixed strict budget ($\epsilon=2.0$). Each variant modifies one component while keeping the training protocol, split, and evaluation procedure unchanged. Table~\ref{tab:ablation} summarizes the resulting impact on ranking and rare-defect detection performance.

Removing class balancing (Oversampling=False) causes a collapse in recall (0.125), consistent with majority-class domination during optimization in rare-event settings. To avoid any data leakage, porous-targeted augmentation and oversampling were applied only within the training split; validation and test sets were kept unchanged and were never augmented or oversampled. Replacing FI-LDP with isotropic perturbation ($\beta=0$) reduces AUPR from 0.664 to 0.621 (a +6.9\% relative gain for anisotropy), indicating that importance-aware allocation is a primary contributor to utility retention under strict privacy. Finally, graph connectivity influences the precision-recall trade-off: a thermal-only graph ($\alpha=0$) increases AUPR but reduces AUC relative to the multimodal construction, suggesting that geometric features complement thermal similarity by improving global ranking robustness. In summary, FI-LDP improves upon Uniform-LDP most clearly in the moderate privacy regime ($\epsilon \approx 2$--$4$), and the mechanism analysis confirms that this gain is driven by importance-aware anisotropic noise allocation rather than uniform perturbation.

\begin{table}[h]
\centering
\caption{Ablation study of framework components at $\epsilon=2.0$.}
\label{tab:ablation}
\resizebox{\columnwidth}{!}{
\begin{tabular}{l c c c c}
\toprule
\textbf{Configuration} & \textbf{Oversampling} & \textbf{AUC} & \textbf{AUPR} & \textbf{Recall} \\
\midrule
Full Framework & True  & \textbf{0.913} & \textbf{0.664} & \textbf{0.762} \\
w/o Oversampling & False & 0.165 & 0.036 & 0.125 \\
w/o Anisotropy ($\beta=0$) & True & 0.884 & 0.621 & 0.711 \\
Thermal-only Graph ($\alpha=0$) & True & 0.666 & 0.638 & 0.650 \\
\bottomrule
\end{tabular}}
\end{table}

\begin{table}[t]
\centering
\caption{Reproducibility checklist and key hyperparameters for FI-LDP-HGAT.}
\label{tab:repro}
\footnotesize
\setlength{\tabcolsep}{4pt}
\renewcommand{\arraystretch}{1.15}
\begin{tabularx}{\columnwidth}{@{} >{\raggedright\arraybackslash}p{1.85cm} >{\raggedright\arraybackslash}X >{\raggedright\arraybackslash}p{3.05cm} @{}}
\toprule
\textbf{Category} & \textbf{Parameter} & \textbf{Value} \\
\midrule
Data / Protocol
& Train/Val/Test split & 0.6 / 0.2 / 0.2 \\
& Seed-averaged evaluation (runs) & 5 \\
& Decision threshold(s) & $t{=}0.5$; $t^*$ tuned on val \\
\midrule
Graph Construction
& Node feature dimension ($d$) & 64 \\
& Connectivity mixing coefficient ($\alpha$) & main; $\alpha{=}0$ (ablation) \\
& Stratified grouping (layer-wise) & enabled \\
\midrule
HGAT Model
& Hidden dimension; attention heads; layers & 64; 4; 2 \\
& Dropout; attention temperature ($\tatt$) & 0.2; 0.1 \\
\midrule
Optimization
& Optimizer & Adam \\
& Learning rate; weight decay & $10^{-3}$; $10^{-4}$ \\
& Batch size; epochs & 64; 25 \\
& Loss; class balancing & Focal-CE; weighted sampler \\
\midrule
FI-LDP (Importance Prior)
& Warmup to estimate importance $\mathbf{q}$ & enabled \\
& Importance temperature ($\beta$) & 0.6 \\
\midrule
Privacy (LDP)
& Privacy parameter $\delta$ & $10^{-5}$ \\
& Privacy budgets reported ($\epsilon$) & $\{0.5,1,2,4,8\}$ \\
& $\ell_2$ clipping (quantile) & 0.95 \\
\bottomrule
\end{tabularx}
\end{table}

To facilitate reproducibility, the model architecture and training hyperparameters are detailed in Table~\ref{tab:repro}.


\section{Discussion}
\label{sec:discussion}

The experimental results support three main findings regarding the interplay between privacy mechanisms, graph-based relational modeling, and defect detection utility in metal AM. We discuss each in turn, followed by a comparison with prior work on the same dataset and directions for future research.

\subsection{Privacy--Utility Degradation Is Not Inevitable}

The central finding of this work is that privacy-induced utility loss under Local Differential Privacy is not a fixed cost, it depends on whether the perturbation mechanism is aligned with the structure of the learned embedding. Across all experiments, FI-LDP provides the clearest benefit in the moderate privacy regime ($\epsilon \approx 2$--$4$), where the privacy constraint is strong enough to distort high-dimensional features but not so strict as to erase all discriminative information. In this context, FI-LDP consistently improves AUPR and recall relative to Uniform-LDP, which is operationally important in rare-defect monitoring where missed detections are costly. The mechanism-level evidence (Fig.~\ref{fig:mechanism_insight}) is consistent with these gains: the heavy-tailed feature-importance profile shows that predictive utility is concentrated in a small subset of coordinates, and FI-LDP explicitly allocates less noise to this high-importance subset while shifting perturbation to the low-importance tail. The strong negative importance--noise coupling (Spearman $\rho=-0.81$) supports the interpretation that the utility improvement is mechanistic and principled rather than an incidental hyperparameter effect.

In contrast, DP-SGD-style training, which clips and perturbs gradients at every optimization step, suffers catastrophic utility collapse (F1$^*<0.21$) even at moderate budgets. This outcome highlights a fundamental distinction between \emph{training-time} and \emph{release-time} privacy: in high-dimensional, severely imbalanced regimes, accumulated gradient noise disrupts the delicate optimization dynamics needed to learn rare-event boundaries. FI-LDP avoids this failure mode by injecting noise once into the learned embedding after training, preserving the optimization trajectory while still providing formal $(\epsilon,\delta)$-LDP guarantees for the released features.

\subsection{System-Level Robustness From Component Interactions}

The ablation study clarifies that robustness under privacy is a system-level outcome arising from the interaction of data balancing, perturbation design, and graph structure. When class balancing is removed, recall collapses to 0.125, indicating that minority-class learning must be preserved during training regardless of the privacy mechanism. Replacing anisotropic perturbation with isotropic noise ($\beta=0$) reduces AUPR from 0.664 to 0.621, confirming that importance-guided allocation is a primary driver of utility retention under strict privacy. The graph connectivity ablation shows that thermal-only connectivity ($\alpha=0$) can improve AUPR by emphasizing local intensity similarity, but the full multimodal graph yields better AUC, suggesting that spatial information stabilizes global ranking across layers and scan tracks when privacy noise perturbs feature geometry. These interactions illustrate that no single component is sufficient; the privacy--utility gains emerge from the coordinated design of all three elements.

\subsection{Non-Private Performance and Comparison with Prior Work}

The non-private baseline comparison (Table~\ref{tab:baseline_nonprivate}) reveals that flat classifiers operating on pre-extracted ResNet-18 image embeddings achieve strong AUC and AUPR values (e.g., SVM: AUC$=0.979$, AUPR$=0.973$). This reflects the discriminative power of the thermal encoder on this dataset and is consistent with prior findings that melt-pool morphology carries strong defect signatures~\cite{zhang2019process,ho2021dlam}. However, these methods use image embeddings alone and do not model relational dependencies. Among graph-based methods, the proposed HGAT achieves the highest calibrated F1$^*$ (0.941), outperforming GCN (0.862) and vanilla GAT (0.854), which tend to over-predict the minority class without edge-affinity priors.

It is also informative to compare with Khanzadeh et al.~\cite{khanzadeh2019situ}, who applied Self-Organizing Map (SOM) clustering on the same LENS Ti-6Al-4V thin-wall dataset. Using a $6\times6$ SOM on spherically transformed thermal distributions, they reported 96.07\% pore detection accuracy, a false alarm rate of 0.128\%, and an F-score of 98.00\% (Table~7 of that study). Several aspects of this comparison merit discussion. First, the SOM approach is an unsupervised anomaly detection method that identifies abnormal melt pools via cluster dissimilarity, whereas the proposed HGAT is a supervised node-level classifier that learns from labeled pore annotations. The two methods address complementary aspects of porosity prediction: SOM detects distributional outliers, while HGAT directly optimizes for defect discrimination under class imbalance. Second, the SOM evaluation uses detection accuracy (fraction of XCT-confirmed pores whose locations overlap with predicted anomalies), which is not directly comparable to the AUC, AUPR, and F1 metrics used in this work. Third, and most important, neither the SOM nor the flat classifiers address the \emph{privacy-constrained} setting that is the focus of this paper. Under any LDP mechanism, the SOM's cluster-based anomaly detection would be disrupted by noise in the thermal features, and the flat classifiers would lose access to clean embeddings. The proposed FI-LDP-HGAT is, to our knowledge, the only method evaluated on this dataset that maintains structured relational inference under formal source-side privacy guarantees.

\subsection{Comparison with Privacy-Preserving Approaches in Manufacturing}

The privacy-aware benchmarking (Table~\ref{tab:privacy_benchmark}) positions FI-LDP relative to alternative protection paradigms. Compared with Bappy et al.~\cite{bappy2025adaptive}, who proposed image-level de-identification (SIA~+~ASIG) for the same DED privacy problem, FI-LDP operates at the embedding level and provides formal $(\epsilon,\delta)$-LDP guarantees rather than heuristic privacy. While direct numerical comparison is not possible due to differences in evaluation protocol and privacy definition, the two approaches are complementary: SIA~+~ASIG masks trajectory information in raw images before encoding, whereas FI-LDP privatizes the encoded features before graph construction. A combined pipeline that applies image-level de-identification followed by FI-LDP at the embedding level could provide defense-in-depth. Compared with model-level perturbation (MNP~\cite{lee2024privacy}) and infrastructure-level protection~\cite{shi2024sensor,oskolkov2025incremental}, FI-LDP addresses a distinct threat model, non--interactive release of learned embeddings, that is not covered by methods protecting model parameters or data in transit (Table~\ref{tab:privacy_comparison}).

\subsection{Future Research Directions}

While this study establishes a robust privacy--utility frontier for experimental data, several directions remain for industrial-scale deployment. First, extending FI-LDP-HGAT to multi-facility \emph{federated learning}~\cite{wang2026privacy} would enable joint training of global quality assurance models without sharing proprietary sensor signatures or facility-specific parameters~\cite{zhou2025privacy}. In such a setting, FI-LDP could serve as the local privatization step within each federated client. Second, future work will investigate \emph{physics-informed graph inductive biases}~\cite{zhou2025spatially}: incorporating explicit process priors (e.g., heat conduction kernels or solidification constraints) into graph construction or message passing may improve robustness under strict privacy noise by anchoring the graph topology to physical invariants rather than noisy feature geometry. Third, a systems-level direction is the use of privatized representations in \emph{autonomous closed-loop control}~\cite{zhang2025advancing}, where FI-LDP embeddings serve as state variables for supervisory decision modules that support real-time defect mitigation, linking privacy-preserving analytics with trustworthy autonomous manufacturing.

\vspace{-6mm}
\section{Conclusion}
\label{sec:conclusion}

This paper introduces FI-LDP-HGAT, a privacy-preserving graph learning framework for in-situ defect monitoring in metal additive manufacturing. The framework addresses a central computational challenge: enabling collaborative analytics from sensitive process data while protecting proprietary information under formal privacy guarantees. The proposed method combines two methodological components---a feature-importance-guided local differential privacy mechanism (FI-LDP) for anisotropic feature privatization, and a stratified Hierarchical Graph Attention Network (HGAT) for physics-informed relational inference---into a coherent pipeline for non-interactive feature release and structure-aware defect prediction.

Experimental evaluation on a DED porosity dataset demonstrates that FI-LDP-HGAT consistently outperforms isotropic privacy baselines and gradient-level privacy approaches across multiple metrics. The method achieves 81.5\% utility recovery at a moderate privacy budget ($\epsilon=4$) and maintains strong defect recall (0.762) under a strict budget ($\epsilon=2$), while DP-SGD-style training collapses entirely under the same constraints. Among non-private baselines, the proposed HGAT achieves the highest calibrated F1$^*$ (0.941), and mechanism-level analysis confirms that the privacy--utility gains of FI-LDP are driven by principled importance-guided noise allocation (Spearman $\rho=-0.81$) rather than incidental effects. These results indicate that anisotropic, importance-guided perturbation can mitigate the utility collapse typically observed in high-dimensional private learning by selectively protecting the most informative feature coordinates. More broadly, this work demonstrates that reliable graph-based defect monitoring and strict local privacy can be reconciled, providing a technically grounded pathway for trustworthy multi-stakeholder AI deployment in metal additive manufacturing.

\vspace{-4mm}

\section*{Competing Interests}
The authors declare that they have no known competing financial interests or personal relationships that could have appeared to influence the work reported in this paper.
\vspace{-2mm}

\section*{Declaration of AI and AI-assisted Technologies in the Writing Process}
During the preparation of this work, the authors used AI assisted tool to refine the linguistic clarity and improve the narrative flow of the manuscript. After using this tool, the authors reviewed and edited the content as needed and take full responsibility for the scientific accuracy and integrity of the final published work.

\bibliographystyle{asmems4}

\bibliography{asme2e}


\end{document}